\documentclass[11pt,a4paper]{article}
\usepackage[hyperref]{acl2021}
\usepackage{times}
\usepackage{latexsym}
\usepackage{amsmath}
\usepackage{booktabs}
\usepackage{subfigure}
\usepackage{graphicx}
\usepackage{makecell}
\usepackage{multirow}

\usepackage{microtype}

\usepackage{enumitem}

\aclfinalcopy 
\def\aclpaperid{1690} 

\title{EarlyBERT: Efficient BERT Training via Early-bird Lottery Tickets}

\author{
Xiaohan Chen$^1$\thanks{~~Work was done when the author interned at Microsoft.} \quad Yu Cheng$^2$ \quad Shuohang Wang$^2$ \quad Zhe Gan$^2$ \\ \textbf{Zhangyang Wang}$^1$  \quad \textbf{Jingjing Liu}$^2$\\ \\
    $^1$University of Texas at Austin, $^2$Microsoft Corporation \\
    {\small \{\tt xiaohan.chen, atlaswang\}@utexas.edu} \\
    \small \{\tt yu.cheng, shuowa, zhe.gan, jingjl\}@microsoft.com
}

\date{}

\begin{document}
\maketitle
\begin{abstract}
Heavily overparameterized language models such as BERT, XLNet and T5 have achieved impressive success 
in many NLP tasks.
However, their high model complexity requires enormous computation resources and extremely long training time for both pre-training and fine-tuning.
Many works have studied model compression on large NLP models, but only focusing on reducing inference time while still requiring an expensive training process. Other works use extremely large batch sizes to shorten the pre-training time, at the expense of higher computational resource demands.
In this paper, inspired by the \textit{Early-Bird Lottery Tickets} recently studied for computer vision tasks, we propose EarlyBERT, a general computationally-efficient training algorithm applicable to both pre-training and fine-tuning of large-scale language models. 
By slimming the self-attention and fully-connected sub-layers inside a transformer, we are the first to identify \emph{structured} winning tickets in the early stage of BERT training. We apply those tickets towards efficient BERT training, and conduct comprehensive pre-training and fine-tuning experiments on GLUE and SQuAD downstream tasks. Our results show that EarlyBERT achieves comparable performance to standard BERT, with 35$\sim$45\% less training time. 
Code is available at \url{https://github.com/VITA-Group/EarlyBERT}.
\end{abstract}

\section{Introduction}
Large-scale pre-trained language models (\emph{e.g.,} BERT~\citep{devlin2018bert}, XLNet~\citep{yang2019xlnet}, T5~\citep{raffel2019exploring}) have significantly advanced the state of the art in the NLP field.
Despite impressive empirical success, their computational inefficiency has become an acute drawback in practice. As more transformer layers are stacked with larger self-attention blocks, model complexity increases rapidly. For example, compared to BERT-Large model with 340 million parameters, T5 has more than 10 billion to learn.
Such high model complexity calls for expensive computational resources and extremely long training time.

Model compression is one approach to alleviating this issue. Recently, many methods have been proposed to encode large NLP models compactly~\citep{sanh2019distilbert,jiao2019tinybert,sun2019patient,mobilebert}. 
However, the focus is solely on reducing inference time or resource costs, leaving the process of searching for the right compact model ever more costly.
Furthermore, most model compression methods start with a large pre-trained model, which may not be available in practice.
Recent work \citep{You2020Large} proposes to use large training batches, which significantly shortens pre-training time of BERT-Large model but demands daunting computing resources (1,024 TPUv3 chips). 

In contrast, our quest is to find a general resource-\emph{efficient} training algorithm for large NLP models, which can be applied to both pre-training and fine-tuning stages. Our goal is to trim down the training time and avoid more costs of the total training resources (\emph{e.g.}, taking large-batch or distributed training). To meet this challenge demand, we draw inspirations from  recent work \citep{You2020Drawing} that explores the use of Lottery Ticket Hypothesis (LTH) for efficient training of computer vision models. LTH was first proposed in \citet{frankle2018the} as an exploration to understand the training process of deep networks. 
The original LTH substantiates a trainable sparse sub-network at initialization, but it cannot be directly utilized for efficient training, since the subnetwork itself has to be searched through a tedious iterative process. In addition, most LTH works discussed only unstructured sparsity.
The study of \citet{You2020Drawing} presents discoveries that structured lottery tickets can emerge in early stage of training (\emph{i.e.}, Early-Bird Ticket), and therefore a structurally sparse sub-network can be identified with much lower costs, leading to practical efficient training algorithms.

Inspired by the success of LTH and Early-Bird Ticket, we propose EarlyBERT, a general efficient training algorithm based on structured Early-Bird Tickets. Due to the vast differences between the architectures and building blocks of computer vision models and BERT, directly extending the method of \citet{You2020Drawing} does not apply to our work.
By instead using network slimming~\citep{liu2017learning} on the self-attention and fully-connected sub-layers inside a transformer, we are the first to introduce an effective approach that can identify \textit{structured winning tickets in the early stage of BERT training}, that are successfully applied for efficient language modeling pre-training and fine-tuning. 
Extensive experiments on BERT demonstrate that EarlyBERT can save 35$\sim$45\% training time with minimal performance degradation, when evaluated on GLUE and SQuAD benchmarks.

\section{Related Work}

\paragraph{Efficient NLP Models}
It is well believed that BERT and other large NLP models are considerably overparameterized \citep{mccarley2019pruning,sun2019patient}. This explains the emergence of many model compression works, which can be roughly categorized into quantization \citep{shen2020q,zafrir2019q8bert}, knowledge distillation \citep{sun2019patient,jiao2019tinybert,sanh2019distilbert,sun2020contrastive,mobilebert}, dynamic routing \citep{fan2019reducing,deebert}, and pruning \citep{li2020train,wang2019structured,mccarley2019pruning,michel19neurips}. Almost all model compression methods focus on reducing inference time, while their common drawback is the reliance on fully-trained and heavily-engineered dense models, before proceeding to their compact, sparse versions - which essentially transplants the resource burden from the inference to the training stage.

Pruning is the mainstream approach for compressing BERT so far \cite{gordon2020compressing}.  \citet{mccarley2019pruning} proposed to greedily and iteratively prune
away attention heads contributing less to the model. \citet{wang2019structured} proposed to structurally prune BERT models using low-rank factorization and augmented Lagrangian $\ell_0$ norm regularization. \citet{mccarley2019pruning} pruned less important self-attention heads and slices of MLP layers by applying $\ell_0$ regularization to the coefficient corresponding to each head/MLP layer. Others aim to reduce the training time of transformer-based models via large-batch training and GPU model parallelism \citep{You2020Large,Shoeybi2019MegatronLMTM}. Our work is orthogonal to these works, and can be readily combined for further efficiency boost.

\vspace{-2mm}
\paragraph{Lottery Ticket Hypothesis in Computer Vision}
Lottery Ticket Hypothesis (LTH) was firstly proposed in \citet{frankle2018the}, which shed light on the existence of sparse sub-networks (\emph{i.e.}, winning tickets) at initialization with non-trivial sparsity ratio that can achieve almost the same performance (compared to the full model) when trained alone. The winning tickets are identified by pruning fully trained networks using the so-called Iterative Magnitude-based Pruning (IMP). However, IMP is expensive due to its iterative nature. Moreover, IMP leads to unstructured sparsity, which is known to be insufficient in reducing training cost or accelerating training speed practically. These barriers prevent LTH from becoming immediately helpful towards efficient training. 

\citet{morcos2019one} studies the transferability of winning tickets between datasets and optimizers. \citet{zhou2019deconstructing} investigates different components in LTH and observes the existence of super-masks in winning tickets. Lately,  \citet{You2020Drawing} pioneers to identify Early-Bird Tickets, which emerge at the early stage of the training process, and contain structured sparsity when pruned with Network Slimming \citep{liu2017learning} which adopts channel pruning. Early-bird tickets mitigate the two limitations of IMP aforementioned, and renders it possible to training deep models efficiently, by drawing tickets early in the training and then focusing on training this compact subnetwork only. \citet{chen2021ultra} reveals the benefit of LTH in data-efficient training, but their focus is not on saving training resources.

\vspace{-2mm}
\paragraph{Lottery Ticket Hypothesis in NLP}
All above works evaluate their methods on computer vision models. For NLP models, previous work has also found that matching subnetworks exist in transformers and LSTMs
\citep{yu2019playing,renda2020comparing}. 
\citet{rigl} derived an algorithm for training sparse neural networks according to LTH and applied it to character-level language modeling on WikiText-103. For BERT models, a latest work \citep{chen2020lottery} found that the pre-trained BERT models contain sparse subnetworks, found by unstructured IMP at 40\% to 90\% sparsity, that are independently trainable and transferable to a range of downstream tasks with no performance degradation. Their follow-up work \citep{chen2020lottery2,gan2021playing} pointed out similar phenomenons in pre-trained computer vision and vision-language models. Another work \citep{prasanna2020bert} aims to find structurally sparse lottery tickets for BERT, by pruning entire attention heads and MLP layers. Their experiments turn out that all subnetworks (``good" and ``bad") have ``comparable performance" when fined-tuned on downstream tasks, leading to their ``all tickets are winning" conclusion. 

Nevertheless, both relevant works \citep{chen2020lottery,prasanna2020bert} examine only the pre-trained BERT model, \emph{i.e.}, finding tickets with regard to the fine-tuning stage on downstream tasks. To our best knowledge, no existing study analyzes the LTH at the pre-training stage of BERT; nor has any work discussed efficient BERT training using LTH, for either pre-training or fine-tuning. Our work makes the first attempt of introducing LTH to both efficient pre-training and efficient fine-tuning of BERT. Our results also provide positive evidence that LTH and Early-Bird Tickets in NLP models are amendable to structured pruning.

\section{The EarlyBERT Framework}

In this section, we first revisit the original Lottery Ticket Hypothesis (LTH) \citep{frankle2018the} and its variant Early-Bird Ticket \citep{You2020Drawing}, then describe our proposed EarlyBERT.

\subsection{Revisiting Lottery Ticket Hypothesis}

Denote $f(x;\theta)$ as a deep network parameterized by $\theta$ and $x$ as its input. A sub-network of $f$ can be characterized by a binary mask $m$, which has exactly the same dimension as $\theta$. When applying the mask $m$ to the network, we obtain 
the sub-network $f(x; \theta\odot m)$, where $\odot$ is the Hadamard product operator.
LTH states that, for a network initialized with $\theta_0$, an algorithm called Iterative Magnitude Pruning (IMP) can identify a mask $m$ such that the sub-network $f(x; \theta_0 \odot m)$ can be trained to have no worse performance than the full model $f$ following the same training protocol.
Such a sub-network $f(x; \theta_0 \odot m)$, including both the mask $m$ and initial parameters $\theta_0$, is called a \textit{winning ticket}.
The IMP algorithm works as follows: $(1)$ initialize $m$ as an all-one mask; $(2)$ fully train $f(x; \theta_0 \odot m)$ to obtain a well-trained $\theta$; $(3)$ remove a small portion of weights with the smallest magnitudes from $\theta\odot m$ and update $m$; $(4)$ repeat $(2)$-$(3)$ until a certain sparsity ratio is achieved.

Two obstacles prevent LTH from being directly applied to efficient training.
First, the iterative process in IMP is essential to preserve the performance of LTH; however, this is computationally expensive, especially when the number of iterations is high.
Second, the original LTH does not pursue any structured sparsity in the winning tickets. In practice, unstructured sparsity is difficult to be utilized for computation acceleration even when the sparsity ratio is high~\citep{wen2016learning}.

To mitigate these gaps, Early-Bird Tickets are proposed by \citet{You2020Drawing}, who discovers that when using structured mask $m$ and a properly selected learning rate, the mask $m$ quickly converges and the corresponding mask emerges as the winning ticket in the early stage of training. The early emergence of winning tickets and the structured sparsity are both helpful in reducing computational cost in the training that follows.
\citet{You2020Drawing} focuses on computer vision tasks with convolutional networks such as VGG \citep{simonyan2014very} and ResNet \citep{he2016deep}.
Inspired by this, we set out to explore whether there are structured winning tickets in the early stage of BERT training that can significantly accelerate language model pre-training and fine-tuning.

\subsection{Discovering EarlyBERT}
\label{subsec:discover-earlybert}

The proposed EarlyBERT\footnote{EarlyBERT refers to the winning ticket discovered by the proposed 3-stage framework, which is equivalent to the resulting pruned BERT model after drawing the winning ticket. We also interchangeably use EarlyBERT as the name of the proposed framework.}
training framework consists of three steps: $(i)$ \emph{Searching Stage}: jointly train BERT and the sparsity-inducing coefficients to be used to draw the winning ticket;
$(ii)$ \emph{Ticket-drawing Stage}: draw the winning ticket using the learned coefficients; and
$(iii)$ \emph{Efficient-training Stage}: train EarlyBERT for pre-training or downstream fine-tuning. 

\vspace{-2mm}
\paragraph{Searching Stage}
To search for the key sub-structure in BERT, we follow the main idea of Network Slimming (NS)  \citep{liu2017learning}. 
However, pruning in NS is based on the scaling factor $\gamma$ in batch normalization, which is not used in most NLP models such as BERT. Therefore, we make necessary modifications to the original NS so that it can be adapted to pruning BERT. Specifically, we propose to associate attention heads and intermediate layers of the fully-connected sub-layers in a transformer with learnable coefficients, which will be jointly trained with BERT but with an additional $\ell_1$ regularization to promote sparsity.

Some studies~\citep{michel19neurips,voita2019analyzing} find that the multi-head self-attention module of transformer can be redundant, presenting the possibility of pruning some heads from each layer of BERT without hurting model capacity. A multi-head attention module \citep{vaswani2017attention} is formulated as:
\begin{align}
    \nonumber
    &\mathrm{MultiHead}(Q,K,V) = \mathrm{Concat}(\mathrm{h}_1,\dots,\mathrm{h}_n) W^O \label{eqn:multihead_att}\\ 
    &\mathrm{h}_i = \mathrm{Attention}(QW^Q_i, KW^K_i, VW^V_i),
\end{align}
where $n$ is the number of heads, and the projections $W^O,W^Q_i,W^K_i,W^V_i$ are used for output, query, key and value. Inspired by \citet{liu2017learning}, we introduce a set of  scalar coefficients $c^h_i$ ($i$ is the index of attention heads and $h$ means ``head'')  inside $\mathrm{h}_i$:
\begin{align}
    \mathrm{h}_i &= c^h_i \cdot \mathrm{Attention}(QW^Q_i, KW^K_i, VW^V_i).
\end{align}

After the self-attention sub-layer in each transformer layer, the output $\mathrm{MultiHead}(Q,K,V)$ will be fed into a two-layer fully-connected network, in which the first layer increases the dimension of the embedding
by 4 times and then reduces it back to the hidden size (768 for BERT\textsubscript{BASE} and 1,024 for BERT\textsubscript{LARGE}). We multiply learnable coefficients to the intermediate neurons:
\begin{align}
    \mathrm{FFN}(x) & = c^f \cdot \mathrm{max}(0, x W_1+b_1) W_2 + b_2.
    \label{eq:bert-fc}
\end{align}
These modifications allow us to jointly train BERT with the coefficients, using the following loss:
\begin{align}
    \mathcal{L}(f(\cdot; \theta), c) = \mathcal{L}_0(f(\cdot; \theta), c) + \lambda\|c\|_1,
\end{align}
where $\mathcal{L}_0$ is the original loss function used in pre-training or fine-tuning, $c$ is the concatenation of all the coefficients in the model including those for attention heads and intermediate neurons, and $\lambda$ is the hyper-parameter that controls the strength of regularization.

Note that in this step, the joint training of BERT and the coefficients are still as expensive as normal BERT training.
However, the \emph{winning strategy} of EarlyBERT is that we only need to perform this joint training for a few steps, before the winning ticket emerges, which is much shorter than the full training process of pre-training or fine-tuning.
In other words, we can identify the winning tickets at a very low cost compared to the full training.
Then, we draw the ticket (\emph{i.e.}, the EarlyBERT), reset the parameters and train EarlyBERT that is computationally efficient thanks to its structured sparsity.
Next, we introduce how we draw EarlyBERT from the learned coefficients.

\vspace{-2mm}
\paragraph{Ticket-drawing Stage}
After training BERT and coefficients $c$ jointly, we draw EarlyBERT using the learned coefficients with a magnitude-based metric.
Note that we prune attention heads and intermediate neurons separately, as they play different roles.

We prune the attention heads whose coefficients have the smallest magnitudes, and remove these heads from the computation graph. We also prune the rows in $W^O$ (see Eqn. (\ref{eqn:multihead_att})) that correspond to the removed heads.
Note that this presents a design choice: should we prune the heads \emph{globally} or \emph{layer-wisely}? 
In this paper, we use layer-wise pruning for attention heads, because the number of heads in each layer is very small (12 for BERT\textsubscript{BASE} and 16 for BERT\textsubscript{LARGE}).
We observe empirically that if pruned globally, the attention heads in some layers may be completely removed, making the network un-trainable. Furthermore, \citet{ramsauer2020hopfield} observes that attention heads in different layers exhibit different behaviors. This also motivates us to only compare importance of attention heads within each layer.

\begin{figure*}[t!]
  \centering
  \begin{tabular}{cc}
    \subfigure[Self-attention in Pre-training]{
        \centering
        \includegraphics[width=.32\textwidth]{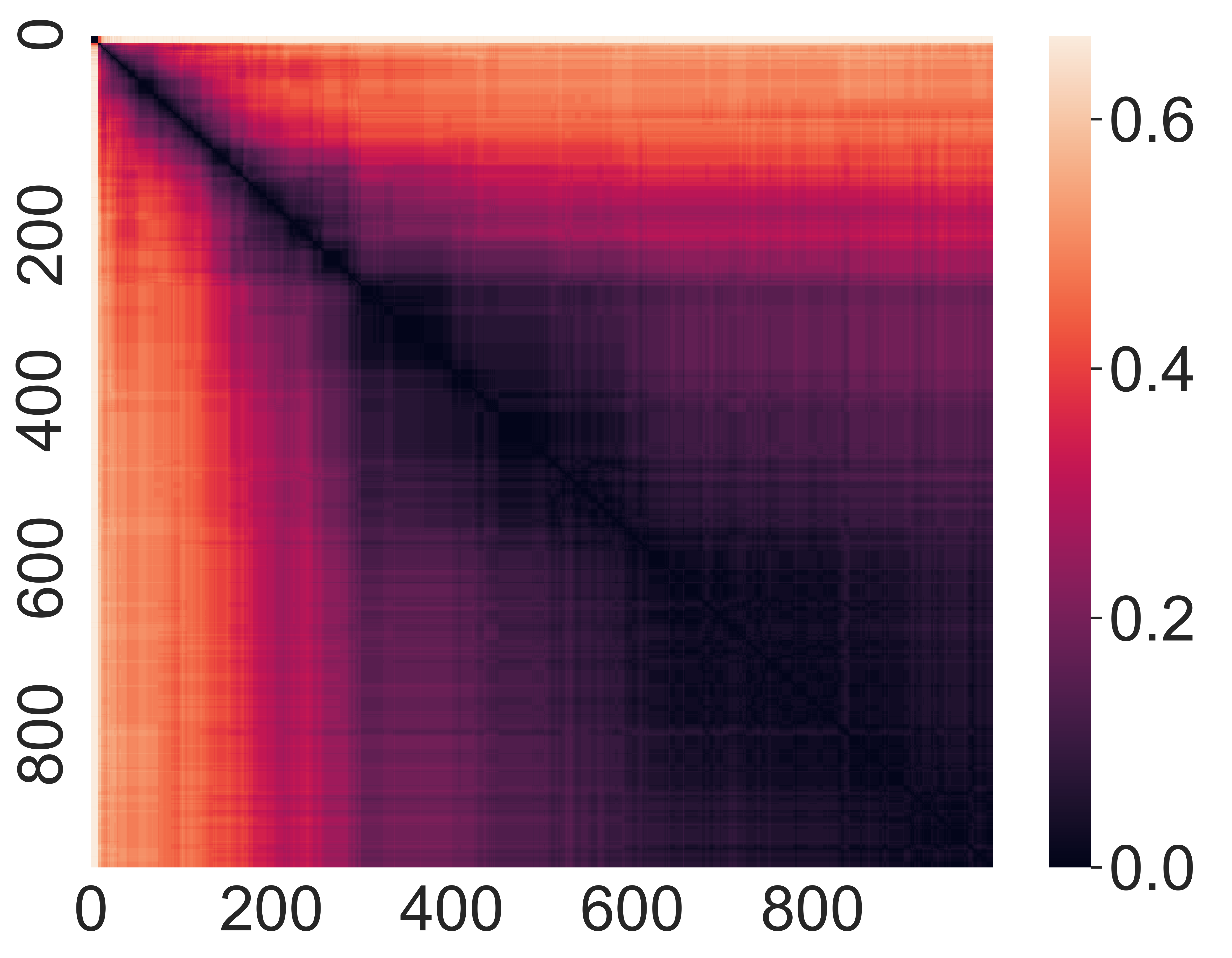}
        \label{sfig:self-pretrain}
    }
    &
    \subfigure[FC in Pre-training]{
        \centering
        \includegraphics[width=.32\textwidth]{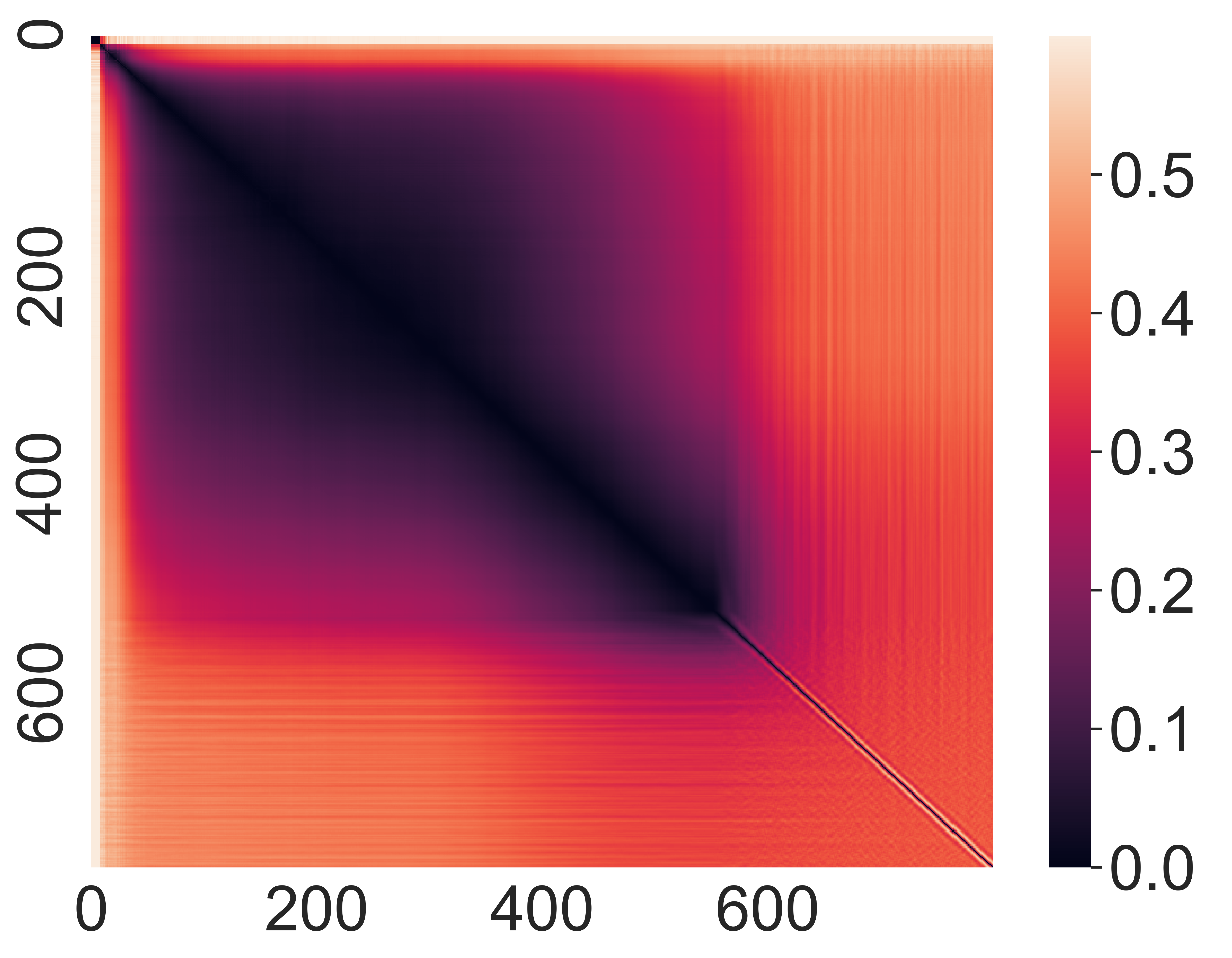}
        \label{sfig:inter-pretrain}
    }
    \\
    \subfigure[Self-attention in Fine-tuning]{
        \includegraphics[width=.32\textwidth]{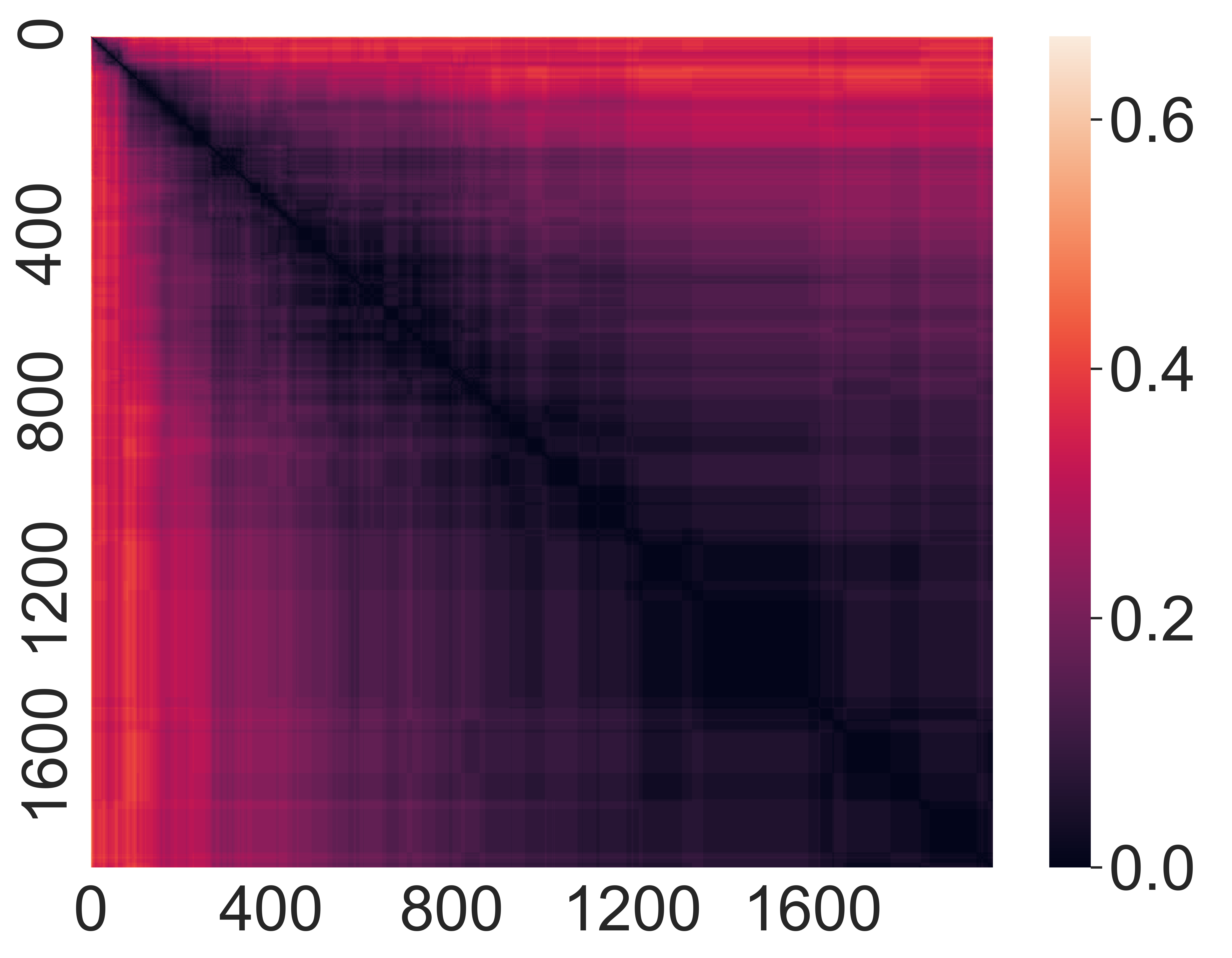}
        \label{sfig:self-finetune}
    }
    &
    \subfigure[FC in Fine-tuning]{
        \includegraphics[width=.32\textwidth]{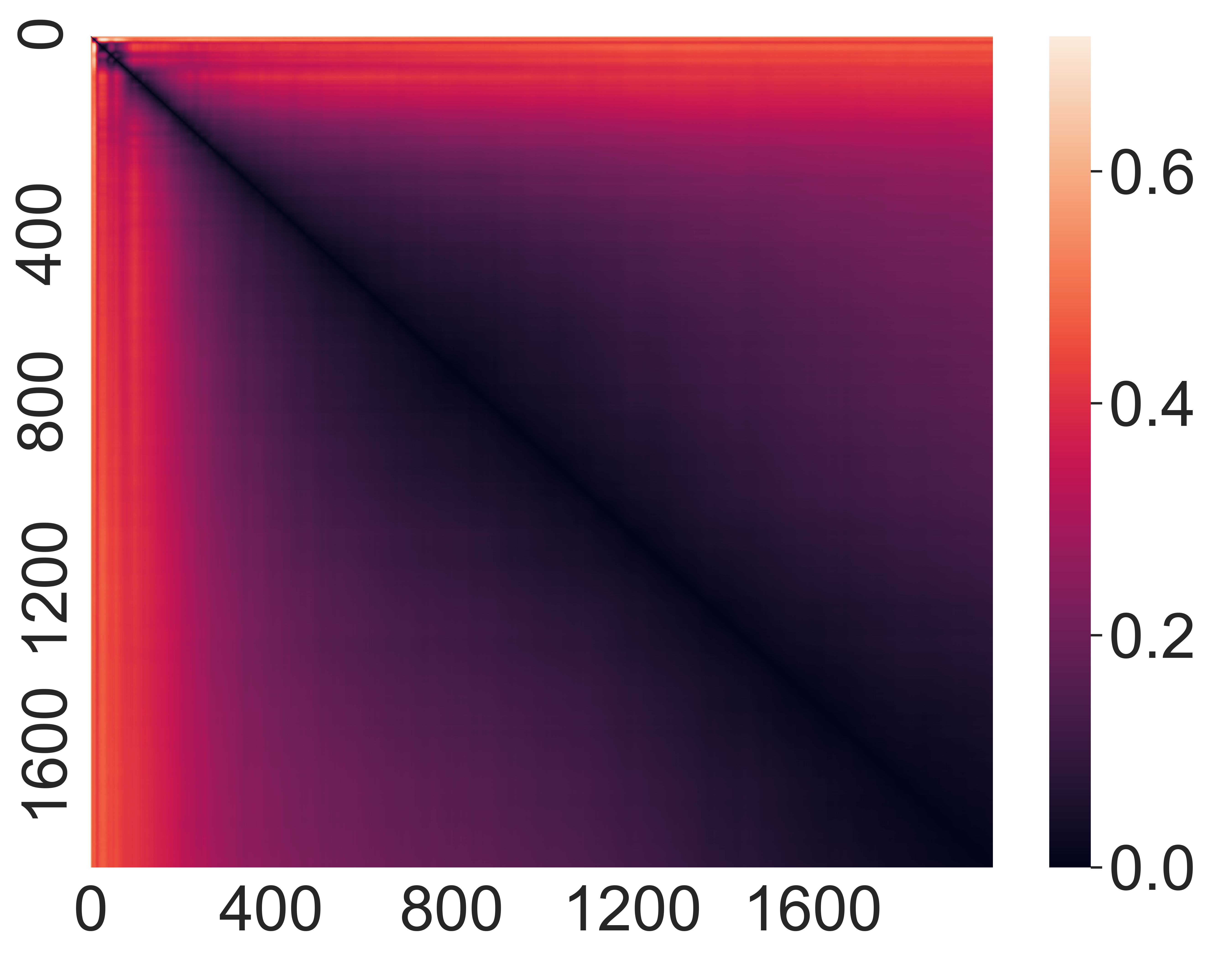}
        \label{sfig:inter-finetune}
    }
    \end{tabular}
    \vspace{-3mm}
  \caption{Illustration of mask difference in Hamming distance. Top: mask distance observed in pre-training. Bottom: mask distance observed in fine-tuning.
  The color represents the normalized mask distance between different training steps. The darker the color, the smaller the mask distance. In both cases, the mask converges quickly, which indicates the early emergence of the tickets.
  }
  \label{fig:mask_distance}
  \vspace{-3mm}
\end{figure*}

Similar to pruning attention heads, we prune intermediate neurons in the fully-connected sub-layers. Pruning neurons is equivalent to reducing the size of intermediate layers, which leads to a reduced size of the weight matrices $W_1$ and $W_2$ in Eqn. (\ref{eq:bert-fc}). Between global and layer-wise pruning, empirical analysis shows that global pruning works better. We also observe that our algorithm naturally prunes more neurons for the later layers than earlier ones, which coincides with many pruning works on vision tasks. We leave the analysis of this phenomenon as future work.

\vspace{-2mm}
\paragraph{Efficient-training Stage}
We then train EarlyBERT that we have drawn for pre-training or fine-tuning depending on the target task.
If we apply EarlyBERT to pre-training, the initialization $\theta_0$ of BERT will be a random initialization, the same setting as the original LTH \citep{frankle2018the} and Early-Bird Tickets \citep{You2020Drawing}.
If we apply EarlyBERT to fine-tuning, then $\theta_0$ can be any pre-trained model.
We can also moderately reduce the training steps in this stage without sacrificing performance, which is empirically supported by the findings in \citet{frankle2018the,You2020Drawing} that the winning tickets can be trained more effectively than the full model.
In practice, the learning rate can also be increased to speed up training, in addition to reducing training steps.

Different from unstructured pruning used in LTH and many other compression works \citep{frankle2018the,chen2020lottery}, structurally pruning attention heads and intermediate neurons in fully-connected layers can directly reduce the number of computations required in the transformer layer, and shrink the matrix size of the corresponding operations, yielding a direct reduction in computation and memory costs.

\subsection{Validation of EarlyBERT}
\label{subsec:earlybert-work}

\paragraph{Early Emergence}
Following a similar manner in \citet{You2020Drawing}, we visualize the normalized mask distance between different training steps, to validate the early emergence of winning tickets. In Figure~\ref{fig:mask_distance}, the axes in the plots are the number of training steps finished.
We only use one fully-connected sub-layer to plot Figure~\ref{sfig:inter-pretrain},\ref{sfig:inter-finetune} due to high dimensionality.
In both pre-training and fine-tuning, the mask converges in a very early stage of the whole training process. Although we observe an increase of mask distance in fully-connected layers during pre-training (in Figure~\ref{sfig:inter-pretrain}), this can be easily eliminated by early stopping and using mask distance as the exit criterion.
An ablation study on how early stopping influences the performance of EarlyBERT is presented in Sec.~\ref{sec:exp-finetuning}.

\begin{table}[t!]
\resizebox{1.0\linewidth}{!}{
\begin{tabular}{lllll}
\toprule
Methods                         & MNLI  & QNLI  & QQP   & SST-2 \\ \midrule
BERT\textsubscript{BASE}        & 83.16 & 90.59 & 90.34 & 91.70 \\
EarlyBERT\textsubscript{BASE}   & 83.58 & 90.33 & 90.41 & 92.09 \\
Random                          & 82.26 & 88.87 & 0.12  & 91.17 \\ \midrule
Methods                         & CoLA  & RTE   & MRPC  &  \\ \midrule
BERT\textsubscript{BASE}        & 0.535 & 65.70 & 80.96 &  \\
EarlyBERT\textsubscript{BASE}   & 0.527 & 66.19 & 81.54 &  \\
Random                          & 0.514 & 63.86 & 78.57 &  \\
\bottomrule
\end{tabular}
}
\vspace{-2mm}
\caption{Comparison between randomly-pruned models and EarlyBERT on GLUE tasks. Different from experiments in Sec.~\ref{sec:exp}, here we prune only 4 heads in each layer and no intermediate neurons.}
\label{tab:exp-random}
\vspace{-4mm}
\end{table}

\begin{table*}[t!]
\centering
\begin{tabular}{lcccccc}
\toprule
Methods                           & MNLI & QNLI & QQP & SST-2 & SQuAD & Time Saved\footnotemark \\ 
\midrule 
BERT\textsubscript{BASE}          & 83.16 & 90.59 & 90.34   & 91.70 & 87.50 & -  \\
EarlyBERT\textsubscript{BASE}     & 81.81 & 89.18 & 90.06 & 90.71 & 86.13 &  40$\sim$45\%  \\
Random\textsubscript{BASE}        & 79.92 & 84.46 & 89.42 & 89.68 & 84.47 & 45$\sim$50\% \\
LayerDrop~\citep{fan2019reducing} &  81.27 & 88.91 & 88.06 & 89.89 & 84.25  & $\sim$33\% \\
\midrule 
BERT\textsubscript{LARGE}         & 86.59 & 92.29 & 91.59 & 92.21 & 90.76 & -  \\
EarlyBERT\textsubscript{LARGE}    & 85.13 & 89.22 & 90.64 & 90.94 & 89.45 & 35$\sim$40\% \\
Random\textsubscript{LARGE}       & 78.45 & 84.46 & 89.89 & 88.65 & 88.79 &  40$\sim$45\% \\
LayerDrop~\citep{fan2019reducing} & 85.12 & 91.12 & 88.88 & 89.97 & 89.44 &  $\sim$33\%\\
\bottomrule
\end{tabular}
\vspace{-2mm}
\caption{Performance of EarlyBERT (fine-tuning) compared with different baselines. We follow the official implementation of LayerDrop method \cite{fan2019reducing}. The protocol that we follow for measuring the training time savings is described in Sec.~\ref{subsec:exp-settings}. We only evaluate models on large downstream tasks since our goal is improving training efficiency.}
\label{tab:exp-finetuning-comparison}
\vspace{-4mm}
\end{table*}

\vspace{-2mm}
\paragraph{Non-trivial Sub-network}
Here, by \textit{non-trivial} we mean that with the same sparsity ratio as in EarlyBERT, randomly pruned model suffers from significant performance drop. The performance drop happens even if we only prune attention heads.
We verify this by running fine-tuning experiments on BERT\textsubscript{BASE}. Specifically, we prune 4 heads from each transformer layer in BERT\textsubscript{BASE} and EarlyBERT.
We fine-tune BERT\textsubscript{BASE} for 3 epochs with an initial learning rate $2\times 10^{-5}$.
We run the searching stage for 0.2 epochs with $\lambda=1\times 10^{-4}$, draw EarlyBERT with pruning ratio $\rho=1/3$, and then fine-tune EarlyBERT for 2 epochs with doubled initial learning rate.
For the randomly pruned models, we randomly prune 4 heads in each layer and follow the same fine-tuning protocol as EarlyBERT. The reported results of randomly pruned models are the average of 5 trials with different seeds for pruning.
The results on four tasks from GLUE benchmark \citep{wang2018glue} presented in Table~\ref{tab:exp-random} show that randomly pruned model consistently under-performs EarlyBERT with a significant gap, supporting our claim that EarlyBERT indeed identifies non-trivial sub-structures.

\section{Experiments}
\label{sec:exp}

\subsection{Experimental Setting}
\label{subsec:exp-settings}

\paragraph{Backbone Models}
Following the official BERT implementation \citep{devlin2018bert,wolf2019huggingface}, we use both BERT\textsubscript{BASE} (12 transformer layers, hidden size 768, 3,072 intermediate neurons, 12 self-attention heads per layer, 110M parameters in total) and BERT\textsubscript{LARGE} (24 transformer layers, hidden size 1,024, 4,096 intermediate neurons, 16 self-attention heads per layer, 340M parameters in total) for experiments.

\vspace{-2mm}
\paragraph{Datasets}
We use English Wikipedia (2,500M words) as the pre-training data. For fine-tuning experiments and evaluation of models in the pre-training experiments, we use tasks from GLUE benchmark \citep{wang2018glue} and a question-answering dataset SQuAD v1.1 \citep{rajpurkar2016squad}.
Note that as our goal is efficient pre-training and fine-tuning, we focus on larger datasets from GLUE (MNLI, QNLI, QQP and SST-2), as it is less meaningful to discuss efficient training on very small datasets.
We use the default training settings for pre-training and fine-tuning on both models. To evaluate model performance, we use Matthew's correlation score for CoLA, matched accuracy for MNLI, F1-score for SQuAD v1.1, and accuracy in percentage for other tasks on GLUE. We omit \% symbols in all the tables on accuracy results.

\vspace{1mm}
\noindent \textbf{Implementation Details}\, For the vanilla BERT, we fine-tune on GLUE datasets for 3 epochs with initial learning rate $2\times 10^{-5}$, and for 2 epochs on SQuAD with initial learning rate $3\times10^{-5}$; we use AdamW \citep{loshchilov2017decoupled} optimizer for both cases. 
For pre-training, we adopt LAMB optimization technique \citep{You2020Large}, which involves two phases of training: the first 9/10 of the total training steps uses a sequence length of 128, while the last 1/10 uses a sequence length of 512. Pre-training by default has 8,601 training steps and uses 64k/32k batch sizes and $6\times10^{-3}$/$4\times10^{-3}$ initial learning rates for the two phases, respectively. All experiments are run on 16 NVIDIA V100 GPUs.

\vspace{1mm}
\noindent \textbf{Training Time Measuring Protocol}\,
We strictly measure the training time saving of EarlyBERT on the QQP task in GLUE using CUDA benchmark mode. To get rid of the influence of the hardware environment at our best, we individually measure the time elapsed during each step and calculate the average time per step over the whole training process. The time for data I/O is excluded. The training time of EarlyBERT includes both the searching stage and the efficient training stage.

\subsection{Experiments on Fine-tuning}
\label{sec:exp-finetuning}

The main results of EarlyBERT in fine-tuning are presented in Table~\ref{tab:exp-finetuning-comparison}.
According to the observation of the early emergence of tickets in Sec.~\ref{subsec:earlybert-work}, we run the searching stage for 0.2 epochs (which accounts for less than 7\% of the cost of a standard 3-epoch fine-tuning) with $\lambda=1\times 10^{-4}$ for all tasks. 
When drawing EarlyBERT, we prune 4 heads in each layer from BERT\textsubscript{BASE} and 6 heads from BERT\textsubscript{LARGE}, \emph{and} globally prune 40\% intermediate neurons in fully-connected sub-layers in both models, 
instead of pruning only heads as in Table~\ref{tab:exp-random}.
After this, we re-train the EarlyBERT models for reduced training epochs (from 3 to 2) on GLUE benchmark and the learning rate scaled up by 2 times to buffer the effect of reduced epochs. For SQuAD dataset, we keep the default setting, as we find SQuAD is more sensitive to the number of training steps.
The selection of these hyperparameters are based on the ablation studies that follow the main results in Table~\ref{tab:exp-finetuning-comparison}, in which we investigate the effects of the number of training epochs, learning rate during downstream fine-tuning, the regularization strength $\lambda$, and the pruning ratios on self-attention heads and intermediate neurons.

\begin{figure}[t!]
    \begin{center}
    \includegraphics[width=0.85\linewidth]{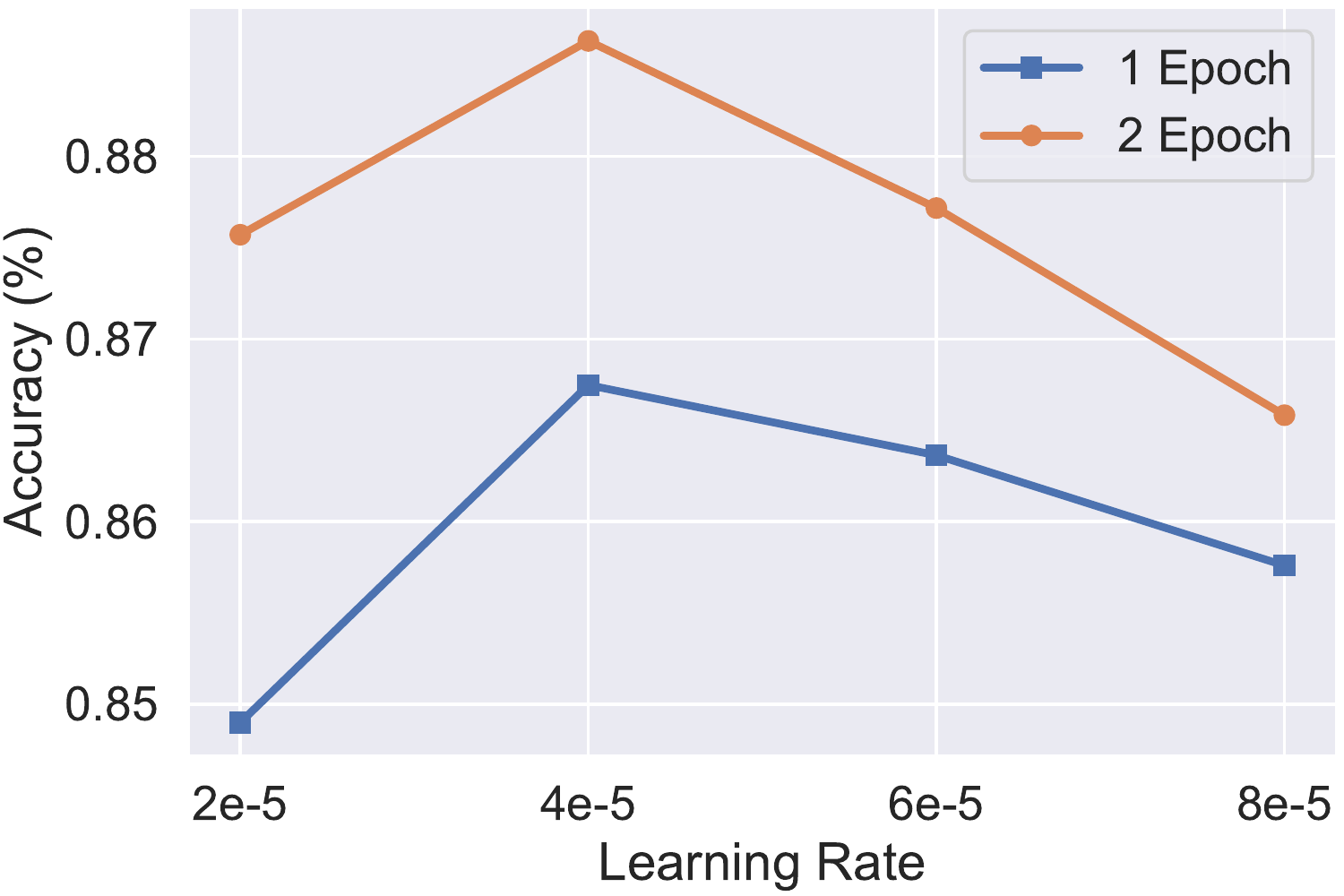}
    \end{center}
    \vspace{-5mm}
    \caption{Effect of reducing training epochs and up-scaling learning rate for EarlyBERT in fine-tuning. The combination of 2-epoch fine-tuning and $4\times10^{-5}$ turns out to be the optimal choice.}
    \label{fig:exp-self-epoch-lr}
    \vspace{-5mm}
\end{figure}

Several observations can be drawn from Table~\ref{tab:exp-finetuning-comparison}.
First, in most tasks, EarlyBERT saves over 40\% of the total training time with 4 self-attention heads pruned in each layer and 40\% FC neurons pruned globally, without inducing much performance degradation. 
Specifically, following the training time measurement protocol in Sec.~\ref{subsec:exp-settings}, we observe that EarlyBERT saves 42.97\% of the total training time of a full BERT model on QQP task. The time saving slightly differs over various tasks, hence we report a range of saving time.
Here, Random\textsubscript{BASE} saves slightly more training time because random pruning skips the searching stage in EarlyBERT\textsubscript{BASE}, but it induces much more accuracy drop.
EarlyBERT\textsubscript{BASE} can also outperform another strong baseline LayerDrop \citep{fan2019reducing}, which drops one third of the layers so that the number of remaining parameters are comparable to ours. Note that LayerDrop models are fine-tuned for three full epochs, yet EarlyBERT is still competitive in most cases.
Second, we consistently observe obvious performance advantage of EarlyBERT over randomly pruned models, which provides another strong evidence that EarlyBERT does discover nontrivial key sparse structures.
Even though there still exists a margin between EarlyBERT and the baseline (\citet{You2020Drawing} also observed similar phenomenon in their tasks), the existence of structured winning tickets and its potential for efficient training is highly promising. We leave as future work to discover winning tickets of higher sparsity but better quality.

\begin{table}[t!]
    \centering
    \begin{tabular}{lccc}
        \toprule
        \hspace{3em}$\lambda$ & $10^{-4}$ & $10^{-3}$ & $10^{-2}$ \\
        \midrule
          & \textbf{88.55}     & 88.43     & 88.42 \\
        \midrule\midrule
        \# Pruned Heads   & 4    & 5    & 6    \\
        \midrule
        Layer-wise pruning   & \textbf{88.55} &  88.13 & 87.65 \\
        \midrule\midrule
        \# Pruned Neurons & 30\% & 40\% & 50\% \\
        \midrule
        Layer-wise pruning       & 88.18 & \textbf{88.22} & 87.90 \\
        Global pruning           & \textbf{88.31} & 88.23 & 87.91
        \\ \bottomrule
    \end{tabular}
    \vspace{-2mm}
    \caption{Ablation of regularization strength $\lambda$ and pruning ratios on self-attention heads and intermediate neurons. All numbers are the average of three runs with different random seeds on four tasks on GLUE (MNLI/QNLI/QQP/SST-2).}
    \label{tab:exp-self-lambda-pruning-ratio}
    \vspace{-5mm}
\end{table}

\begin{table*}[!t]
\centering
\begin{tabular}{lcccccccc}
\toprule
Methods          & CoLA & MNLI  & MRPC  & QNLI  & QQP   & RTE   & SST-2 & SQuAD \\
\midrule
BERT\textsubscript{BASE}       
                 & 0.45 & 81.40 & 84.07 & 89.86 & 89.80 & 60.29 & 90.48 & 87.60 \\
EarlyBERT\textsubscript{BASE}  
                 & 0.41 & 79.97 & 80.39 & 89.86 & 89.44 & 61.01 & 90.94 & 85.48 \\
\midrule
BERT\textsubscript{LARGE}      
                 & 0.50 & 83.56 & 85.90 & 90.44 & 90.45 & 59.93 & 92.55 & 90.43 \\
EarlyBERT\textsubscript{LARGE} 
                 & 0.47 & 82.54 & 85.54 & 90.46 & 90.38 & 61.73 & 91.51 & 89.36
\\ \bottomrule
\end{tabular}
\vspace{-2mm}
\caption{Performance of EarlyBERT (pre-training) compared with BERT baselines. Different from fine-tuning experiments, we evaluate pre-trained models on all downstream tasks in GLUE and SQuAD since fine-tuning cost is negligible compared to the dominant pre-training cost.}
\label{tab:exp-pretrain-ultimate-earlybert}
\vspace{-3mm}
\end{table*}

\begin{figure}[!t]
    \centering
    \includegraphics[width=0.90\linewidth]{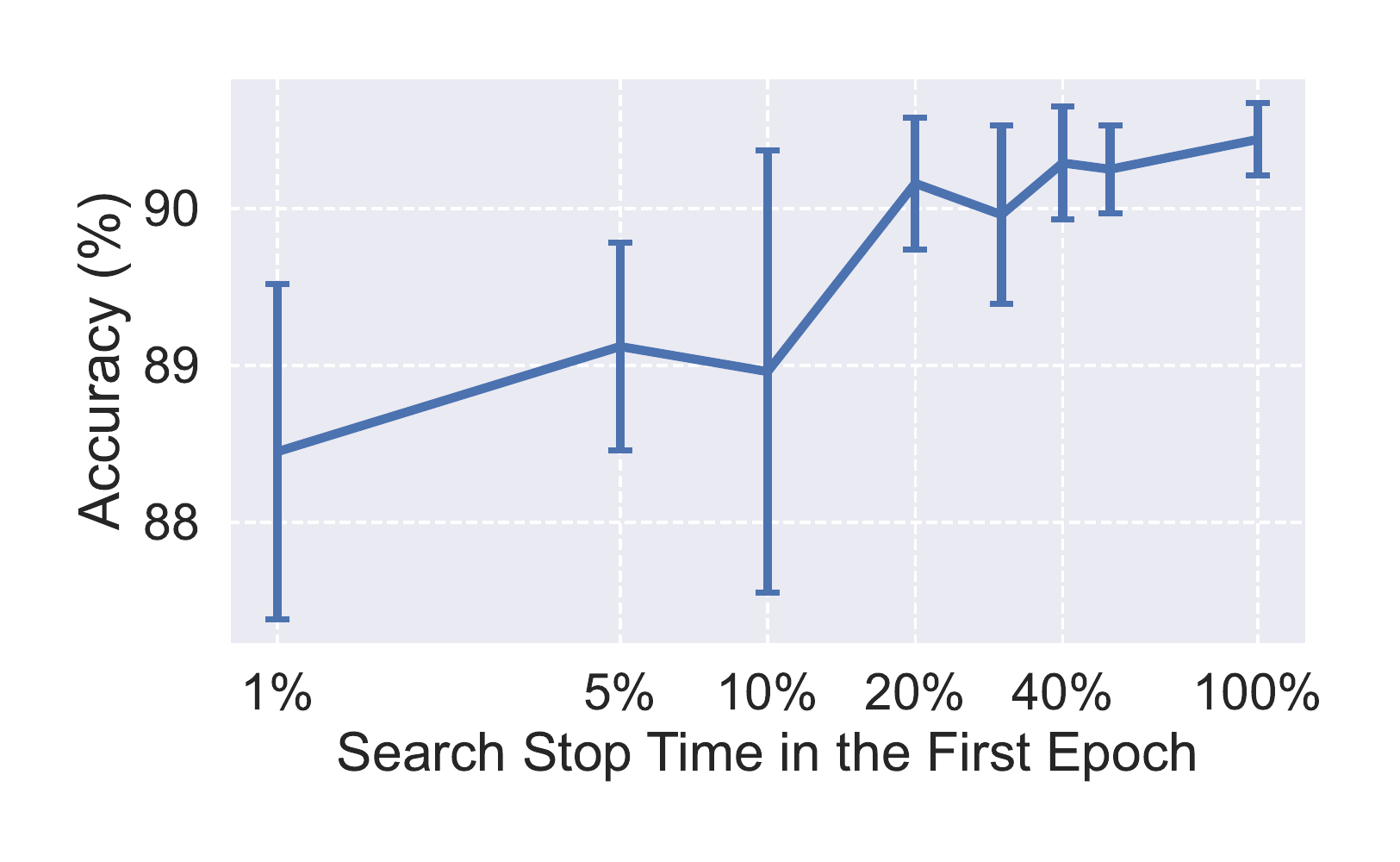}
    \vspace{-1.6em}
    \caption{How various time points of early stopping for searching influences EarlyBERT performance.}
    \label{fig:early-stop-ablation}
    \vspace{-0.6em}
\end{figure}

\vspace{1mm}
\noindent \textbf{Ablation Studies on Fine-tuning}\,
We perform extensive ablation studies to investigate important hyper-parameter settings in EarlyBERT, using
EarlyBERT\textsubscript{BASE} as our testing bed.
For all experiments, we use the average accuracy on the larger datasets from GLUE benchmark (MNLI, QNLI, QQP and SST-2) as the evaluation metric.
\begin{itemize}[leftmargin=*]
    \item \textbf{Number of training epochs and learning rate.} We first investigate whether we can properly reduce the number of training epochs, and if scaling the learning rate can help compliment the negative effect caused by reducing training steps. Results in Figure~\ref{fig:exp-self-epoch-lr} show that when we fine-tune EarlyBERT for fewer epochs on GLUE, up-scaling learning rate first helps to recover performance, and then causes decrease again. We will use two epochs and $4\times10^{-5}$ as learning rate for EarlyBERT on GLUE experiments.
    \vspace{-2mm}
    \item \textbf{Regularization strength $\lambda$.} A proper selection of the regularization strength $\lambda$ decides the quality of the winning ticket, consequently the performance of EarlyBERT after pre-training/fine-tuning. Results in Table~\ref{tab:exp-self-lambda-pruning-ratio} show that $\lambda$ has marginal influence on EarlyBERT performance. We use $\lambda=10^{-4}$ that achieves the best performance in following experiments.
    \vspace{-2mm}
    \item \textbf{Pruning ratios $\rho$.} We further investigate the effects of different pruning ratios as well as layer-wise/global pruning on the performance of EarlyBERT. As discussed in Sec.~\ref{subsec:discover-earlybert}, we only consider layer-wise pruning for self-attention heads. Table~\ref{tab:exp-self-lambda-pruning-ratio} shows that the performance monotonically decreases when we prune more self-attention heads from BERT; however, we see a slight increase and then a sharp decrease in accuracy, when the pruning ratio is raised for intermediate neurons in fully-connected sub-layers ($40\%$ pruning ratio seems to be the sweet spot). We also observe consistent superiority of global pruning over layer-wise pruning for intermediate neurons.\vspace{-2mm}
    \item 
    \textbf{Early-stop strategy for searching.} In Figure~\ref{fig:mask_distance}, we show the early emergence of winning tickets in BERT when trained with $\ell_1$ regularization, suggesting we stop the searching stage early to save computation while still generating high-quality tickets. Here, we study how the early-stop strategy influences the model performance. We fine-tune EarlyBERT on QNLI following the same setting described earlier in this section, but stop the searching stage at different time points during the first epoch for searching.
    Results in Figure~\ref{fig:early-stop-ablation} show ($i$) an abrupt increase in accuracy when we stop at 20\% of the first epoch; ($ii$) slight increase when we delay the stop till the end of the first epoch. Considering training efficiency, we think 20$\sim$40\% makes suitable stopping time.
\end{itemize}

\begin{table}[t!]
\centering
\resizebox{1.0\linewidth}{!}{
\begin{tabular}{@{}ccccc@{}}
\toprule
\multicolumn{1}{c}{Time Saving}
& \multicolumn{1}{c}{\multirow{2}{*}{3 Heads}}
& \multicolumn{1}{c}{\multirow{2}{*}{4 Heads}}
& \multicolumn{1}{c}{\multirow{2}{*}{5 Heads}}
& \multicolumn{1}{c}{\multirow{2}{*}{6 Heads}} \\ \cmidrule(lr){1-1}
\multicolumn{1}{c}{Prune Ratio} & & & & \\ \midrule
\multirow{2}{*}{FC - 30\%} & -35.78\% & -38.66\% & -41.26\% & -45.34\% \\ \cmidrule(lr){2-5}
                           & 89.62\%  & 89.55\%  & 89.60\%  & 89.50\% \\ \midrule
\multirow{2}{*}{FC - 40\%} & -39.72\% & -42.97\% & -43.93\% & -44.49\% \\ \cmidrule(lr){2-5}
                           & 89.66\%  & 89.61\%  & 89.58\%  & 89.38\% \\ \midrule
\multirow{2}{*}{FC - 50\%} & -43.89\% & -45.54\% & -47.02\% & -48.53\% \\ \cmidrule(lr){2-5}
                           & 89.54\%  & 89.35\%  & 89.34\%  & 89.31\% \\
\bottomrule
\end{tabular}
}
\vspace{-0.5em}
\caption{Training time savings vs. accuracies of EarlyBERT on the QQP task in GLUE with different pruning ratios for self-attention heads and FC neurons.}
\label{tab:appendix_2}
\vspace{-0.5em}
\end{table}

\vspace{1mm}
\noindent\textbf{Trade-off Between Efficiency and Performance}
We vary the pruning ratios for the FC layers and the number of self-attention heads pruned in each layer in EarlyBERT, fine-tune the models on QQP in GLUE, and obtain the corresponding validation accuracies and training time savings following the protocol above. Results are shown in Table
\ref{tab:appendix_2}. We can see clear correlations between the training time saving and the accuracy --- the more FC neurons or self-attention heads pruned, the more training time saving yet the larger accuracy drop. Moreover, for most combinations of these two hyper-parameters, the accuracy drop is within 1\%, which also supports the efficiency of EarlyBERT.

\subsection{Experiments on Pre-training}

\begin{figure}[t!]
    \centering
    \includegraphics[width=0.8\linewidth]{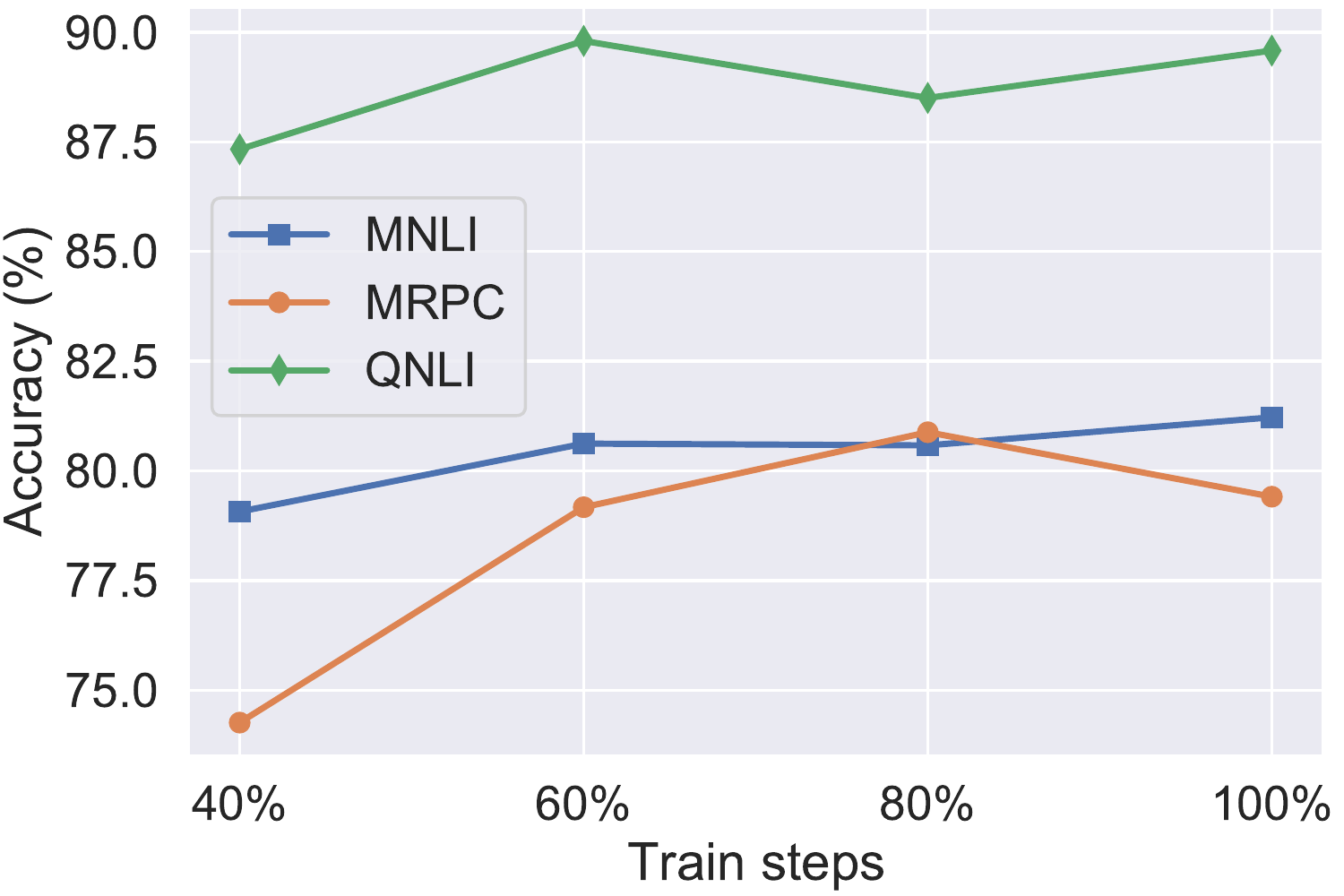}
    \caption{Effect of reducing training steps in pre-training on EarlyBERT\textsubscript{BASE}.}
    \label{fig:exp-self-train-steps}
    \vspace{-4mm}
\end{figure}

We also conduct pre-training experiments and present the main results in Table~\ref{tab:exp-pretrain-ultimate-earlybert}.
We run the search stage for 400 steps of training in the first training phase that uses a sequence length of 128 which only accounts for less than 3\% of a standard pre-training, with $\lambda=1\times 10^{-4}$.
When we draw EarlyBERT, similar to the settings in fine-tuning experiments, we prune 4 heads in each layer from BERT\textsubscript{BASE} and 6 heads from BERT\textsubscript{LARGE}; however, we prune slightly fewer (30\%) intermediate neurons in fully-connected sub-layers in both models, since we empirically observe that pre-training is more sensitive to aggressive intermediate neuron pruning.
In both phases of pre-training, we reduce the training steps to 80\% of the default setting when training EarlyBERT (based on the ablation study shown in Figure~\ref{fig:exp-self-train-steps}).
Other hyper-parameters for pre-training follow the default setting described in Sec.~\ref{subsec:exp-settings}.
All models are fine-tuned and evaluated on GLUE 
and SQuAD v1.1 
with the default setting. 

Different from fine-tuning experiments, the pre-training stage dominates the training time over the downstream fine-tuning, and thus we only consider the training time saving during pre-training.
Since the randomly pruned models do not have competitive performance in fine-tuning experiments 
as shown in Sec.~\ref{sec:exp-finetuning}, 
we focus on comparing 
EarlyBERT with the full BERT baseline.

From the results presented in Table~\ref{tab:exp-pretrain-ultimate-earlybert}, we can see that on downstream tasks with larger datasets such as QNLI, QQP and SST-2, we can achieve accuracies that are close to BERT baseline (within 1\% accuracy gaps except for EarlyBERT\textsubscript{BASE} on MNLI and SQuAD). However, on downstream tasks with smaller datasets, the patterns are not consistent: we observe big drops on CoLA and MRPC but improvement on RTE. Overall, EarlyBERT achieves comparable performance while saving 30$\sim$35\% training time thanks to its structured sparsity and reduction in training steps.

\paragraph{Reducing Training Steps in Pre-training}
We investigate whether EarlyBERT, when non-essential heads and/or intermediate neurons are pruned, can train more efficiently, and whether we can reduce the number of training steps in pre-training.
This can further help reduce training cost in addition to the efficiency gain from pruning.
We use EarlyBERT\textsubscript{BASE}-Self (only self-attention heads are pruned when drawing the winning ticket) as the testing bed.
Figure~\ref{fig:exp-self-train-steps} shows the performance decreases more when we reduce the number of training steps to 60\% or less.
Reducing it to 80\% seems to be a sweet point with the best balance between performance and efficiency.

\vspace{-0.2em}
\subsection{Comparison with Previous Lottery Tickets Work in NLP}
\vspace{-0.2em}

On one hand, two relevant works \cite{chen2020lottery,prasanna2020bert} only investigate lottery tickets on pre-trained NLP models for fine-tuning on the downstream tasks, while EarlyBERT makes the first attempt of introducing lottery tickets to both fine-tuning and pre-training stages, and provides empirical evidence that NLP models are amendable to structured pruning.

On the other hand, EarlyBERT pursues structured sparsity while \citet{chen2020lottery} promotes unstructured sparsity, which is hardware unfriendly and provides almost no acceleration, besides the high cost of IMP. As an implicit comparison, \citet{chen2020lottery} induces 0.4\% accuracy drop on SQuAD v1 dataset compared to the BERT baseline with 40\% \underline{unstructured} sparsity (comparable with our settings in Section~\ref{sec:exp-finetuning}), while EarlyBERT induces 1.37\% accuracy drop. Note that \citet{chen2020lottery} uses 6x training times (because IMP reaches 40\% sparsity with 6 iterations) and 4.69x FLOPs, but EarlyBERT uses only 0.76x training times and FLOPs in contrast.

\section{Conclusion} 
In this paper, we present EarlyBERT, an efficient framework for large-scale language model pre-training and fine-tuning. Based on Lottery Ticket Hypothesis, EarlyBERT identifies structured winning tickets in an early stage, then uses the pruned network for efficient training. Experimental results 
demonstrate that the proposed method is able to achieve comparable performance to standard BERT with much less training time. Future work includes 
exploring more data-efficient strategies to enhance the current training pipeline.

\bibliographystyle{acl_natbib}
\bibliography{anthology,acl2021}


\clearpage

\appendix

\section{More Comparison with BERT Baseline}

For more explicit comparison, we conduct a two-way fine-tuning experiment in addition to the main results in Table~\ref{tab:exp-finetuning-comparison}. All results are averages of 3 runs.

We first increase the training cost of EarlyBERT to match BERT performance by extending the searching stage to a full epoch, which, according to our ablation study in Figure~\ref{fig:early-stop-ablation}, helps to improve the performance of EarlyBERT. In this case, EarlyBERT still has 16\% time and FLOPs savings, with comparable performance shown in Table~\ref{tab:exp-double-axis-same-acc}.

Secondly, we reduce the training steps of BERT to match the FLOPs of EarlyBERT, inducing obvious gaps between BERT and EarlyBERT as presented in Table~\ref{tab:exp-double-axis-same-comp}.

\begin{table}[h]
\resizebox{1.0\linewidth}{!}{
\begin{tabular}{@{}lcccc@{}}
\toprule
\textbf{GLUE Task} & \textbf{MNLI} & \textbf{QNLI} & \textbf{QQP} & \textbf{SST-2} \\ \midrule
BERT               & 83.48\%       & 90.43\%       & 90.37\%      & 91.86\% \\
EarlyBERT          & 83.36\%       & 90.44\%       & 90.33\%      & 91.55\% \\ \bottomrule
\end{tabular}
}
\caption{We increase the number of training steps of EarlyBERT so that it achieves very close performances to the BERT baseline on the larger four tasks in GLUE benchmark.} 
\label{tab:exp-double-axis-same-acc}
\end{table}

\begin{table}[h]
\resizebox{1.0\linewidth}{!}{
\begin{tabular}{@{}lcccc@{}}
\toprule
\textbf{GLUE Task}         & \textbf{MNLI} & \textbf{QNLI} & \textbf{QQP} & \textbf{SST-2} \\ \midrule
\makecell{BERT -\\Reduced} & 82.85\%       & 89.86\%       & 89.45\%      & 91.70\% \\
EarlyBERT                  & 83.26\%       & 90.16\%       & 90.22\%      & 91.67\% \\ \bottomrule
\end{tabular}
}
\caption{Comparison between the performance of BERT and EarlyBERT with the same training time on the larger four tasks on GLUE benchmark by reducing the number of training steps of BERT. Obvious gaps can be observed on all four tasks but SST-2.}
\label{tab:exp-double-axis-same-comp}
\end{table}

\section{Searching EarlyBERT using on the Masked Language Modeling Task}

It is found in \citet{chen2020lottery} selecting a winning ticket for BERT fine-tuning on the masked language modeling task (MLM), i.e., pre-training objective makes for better tickets performing on many of the downstream tasks. Here, we try the experiments of using the MLM objective during the searching stage. Results are summarized in Table~\ref{tab:appendix_3}. Our main observations include:
\begin{itemize}
    \item When using the MLM objective for the searching stage, the mask distance for both self-attention heads and FC neurons converged well and quickly within 100 training steps.
    \item We first apply the global pruning method to the FC neurons because we observed better performance of EarlyBERT with that method. However, while we previously found in EarlyBERT that the latter layers will be pruned more, we observed the opposite phenomenon when using MLM objective --- the former layers are pruned more instead. In terms of accuracy, we observed significant gaps compared to EarlyBERT.
    \item Based on the above observations, we also applied layerwise pruning for MLM experiments (shown in the last row of Table \ref{tab:appendix_3}). We did see improved accuracy with layerwise pruning but the gaps between EarlyBERT are still large (except on QQP).
\end{itemize}

\begin{table}[t!]
\centering
\resizebox{1.0\linewidth}{!}{
\begin{tabular}{@{}lcccc@{}}
\toprule
\textbf{GLUE Task} & \textbf{MNLI} & \textbf{QNLI} & \textbf{QQP} & \textbf{SST-2} \\ \midrule
BERT               & 83.36\%       & 90.53\%       & 90.41\%      & 91.61\%        \\
EarlyBERT          & 81.97\%       & 88.68\%       & 89.26\%      & 90.48\%        \\
MLM - FC Global    & 78.36\%       & 84.84\%       & 88.86\%      & 88.65\%        \\
MLM - FC Layerwise & 79.01\%       & 86.55\%       & 89.16\%      & 88.53\%        \\ \bottomrule
\end{tabular}
}
\caption{Comparison of the accuracies of EarlyBERT and EarlyBERT with winning tickets searched using MLM objective on downstream tasks in GLUE.}
\label{tab:appendix_3}
\end{table}

\begin{table}[b!]
\centering
\begin{tabular}{@{}ccc@{}}
\toprule
\textbf{Training Steps} & \textbf{BERT} & \textbf{EarlyBERT} \\ \midrule
100\%                   & 0.00\%        & 0.00\%             \\
80\%                    & -1.94\%       & +1.96\%            \\
60\%                    & -2.48\%       & -1.42\%            \\
40\%                    & -3.62\%       & -3.43\%            \\ \bottomrule
\end{tabular}
\caption{Effects of reduced training steps for BERT and EarlyBERT in average on GLUE benchmark tasks and SQuAD.}
\label{tab:appendix_4}
\end{table}

\section{The Effect of Reduced Training Steps during Pre-training}

We perform the same as the analysis of the effect of reduced training steps during pre-training in Figure~\ref{fig:exp-self-train-steps} for both the vanilla BERT and EarlyBERT. We calculate how performance will be influenced due to the reduced training steps. We use F1 score for SQuAD, Matthew’s correlation score for CoLA and accuracy for all other tasks on GLUE as the metric. We report the performance reduction (or gain) in percentage average on all tasks, normalized by the performance of baseline, i.e., BERT or EarlyBERT trained with the default number of training steps. Similar metric is used in DistilBERT \citep{sanh2019distilbert}. Results are shown in Table \ref{tab:appendix_4}. We can see that using only 80\% training steps actually improves the performance of EarlyBERT on average but in contrast hurts the performance of BERT. Similarly, using 60\% training steps hurts BERT more than EarlyBERT. And as expected, saving more training steps generally hurt more. We think this is one piece of evidence that motivated us to use reduced training steps for EarlyBERT.


\end{document}